\documentclass[11pt,a4paper]{article}

\usepackage[T1]{fontenc}
\usepackage[utf8]{inputenc}
\usepackage{CJKutf8}
\usepackage{times}
\usepackage{marvosym}

\usepackage[margin=2.4cm,headheight=38pt,headsep=18pt,footskip=28pt]{geometry}
\usepackage{setspace}
\usepackage[table,dvipsnames]{xcolor}
\usepackage{colortbl}
\usepackage{graphicx}
\usepackage{booktabs}
\usepackage{array}
\usepackage{multirow}
\usepackage{tabularx}

\usepackage{amsmath,amssymb}

\definecolor{AMapBlue}{HTML}{48B8F0}
\definecolor{AMapDeepBlue}{HTML}{0877C9}
\definecolor{AMapSky}{HTML}{EAF8FF}
\definecolor{AMapLine}{HTML}{9EDBFA}
\definecolor{TextBlack}{HTML}{20242A}

\usepackage[numbers,sort&compress]{natbib}

\usepackage{enumitem}
\setlist{nosep,leftmargin=1.5em}
\usepackage{caption}
\usepackage{fancyhdr}
\usepackage{lastpage}
\usepackage{titlesec}
\usepackage[most]{tcolorbox}
\usepackage{mdframed}
\usepackage[colorlinks=true,linkcolor=AMapDeepBlue,citecolor=AMapDeepBlue,urlcolor=AMapDeepBlue]{hyperref}

\newcommand{\reportname}{FantasyEvo}
\newcommand{\reportlogo}{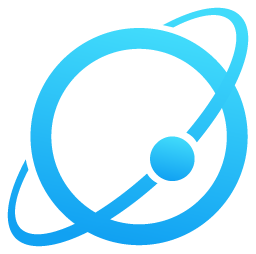}
\newcommand{\reportlogoheight}{18pt}

\newcommand{\orgname}{AMAP CV Lab \textbar{} Alibaba Group}
\newcommand{\reportcenter}[1]{\raisebox{\dimexpr(\depth-\height)/2\relax}{#1}}
\newlength{\reporttextbaselineoffset}
\newcommand{\reporttextalign}[1]{\raisebox{\reporttextbaselineoffset}{#1}}

\renewcommand{\headrulewidth}{1.2pt}
\renewcommand{\footrulewidth}{0.8pt}
\renewcommand{\headrule}{\hbox to\headwidth{\color{AMapBlue}\leaders\hrule height \headrulewidth\hfill}}
\renewcommand{\footrule}{\hbox to\headwidth{\color{AMapLine}\leaders\hrule height \footrulewidth\hfill}}

\titleformat{\section}
  {\Large\bfseries\color{AMapDeepBlue}}
  {\thesection}{0.75em}{}
  [\vspace{0.32em}{\color{AMapLine}\titlerule[0.6pt]}]
\titleformat{\subsection}
  {\large\bfseries\color{AMapDeepBlue}}
  {\thesubsection}{0.65em}{}
\titleformat{\subsubsection}
  {\normalsize\bfseries\color{AMapDeepBlue}}
  {\thesubsubsection}{0.55em}{}
\titlespacing*{\section}{0pt}{1.55em plus 0.25em minus 0.1em}{0.95em}
\titlespacing*{\subsection}{0pt}{1.25em plus 0.2em minus 0.1em}{0.55em}
\titlespacing*{\subsubsection}{0pt}{1.05em plus 0.15em minus 0.08em}{0.45em}

\newtcolorbox{abstractbox}{
  enhanced,
  breakable,
  colback=AMapSky,
  colframe=AMapBlue,
  coltitle=white,
  colbacktitle=AMapBlue,
  title=Abstract,
  fonttitle=\Large\bfseries,
  halign title=center,
  toptitle=5pt,
  bottomtitle=5pt,
  fontupper=\large,
  boxrule=0.8pt,
  arc=1.5mm,
  left=12pt,
  right=12pt,
  top=10pt,
  bottom=10pt,
  before skip=1.2em,
  after skip=1.4em
}

\newmdenv[
  backgroundcolor=gray!8,
  linewidth=0pt,
  innerleftmargin=10pt,
  innerrightmargin=10pt,
  innertopmargin=5pt,
  innerbottommargin=5pt,
  skipabove=5pt,
  skipbelow=5pt
]{cnblock}

\definecolor{DimAction}{HTML}{0877C9}
\definecolor{DimVisual}{HTML}{2E9E5B}
\definecolor{DimPhysics}{HTML}{E07B00}
\definecolor{DimMemory}{HTML}{8E44AD}
\usepackage{pifont}
\newcommand{\mail}{\textsuperscript{\ding{41}}}
\begin{document}
\thispagestyle{fancy}

\begin{center}
  {\bfseries\color{AMapDeepBlue}\LARGE FantasyEvo: Self-Evolving Harnesses for Physically Faithful Video Generation}

\vspace{0.2em}
{\large
  Mingju~Gao\textsuperscript{1,2,*},
  Zhaoxu~Sun\textsuperscript{2,*},
  Fan~Jiang\textsuperscript{2,*,$\dagger$,\mail},
  Ziyu~Zhu\textsuperscript{2},\\
  Mengchao~Wang\textsuperscript{2}, Chiyu~Wang\textsuperscript{2}, Qiuyu~Ding\textsuperscript{2}, Mu~Xu\textsuperscript{2},
  Hao~Tang\textsuperscript{1,\mail}\par
}

\vspace{0.2em}
{\small
  \textsuperscript{*}Equal contribution \quad
  \textsuperscript{$\dagger$}Project Lead \quad
  \textsuperscript{\mail}Corresponding author\par
}

\vspace{0.25em}
{\small
  \textsuperscript{1}School of Computer Science, Peking University \quad
  \textsuperscript{2}AMAP, Alibaba Group\par
}

  \vspace{-0.1em}
\end{center}

\vspace{-1em}

\begin{abstractbox}

Video diffusion models (VDMs) generate visually convincing footage but routinely violate physical laws. Pairing them with reasoning agents that rewrite the prompt seems a natural fix, but existing physics-aware agents gain little: across three VDMs and two benchmarks, the best of them exceeds a hand-written rewriting template by at most 4.6 points and falls below it in three of six settings. 
We argue that what is missing is not more test-time reasoning but better communication with the VDM: the same agent produces markedly different videos depending on how it writes the prompt, and what works for one VDM need not work for another. 
Therefore, we propose FantasyEvo, which learns a model-specific communication policy through iterative self-improvement: a judge agent scores generated videos against physical criteria, an evolution agent revises the policy based on which criteria failed and how the prompt conveyed them, and only revisions that improve held-out performance are kept. 
At test time, the frozen policy adds a single prompt-writing step per task, with one VDM call and no feedback loop. 
On Physics-IQ and StEvo across three VDMs, FantasyEvo outperforms raw prompting by 8.5 and 24.7 points on average, surpassing prior agent-based methods by 3.4 and 13.3 points. 

\end{abstractbox}

\noindent\begin{minipage}{\linewidth}
  \centering
  \captionsetup{type=figure}
  \includegraphics[width=\linewidth,trim={0bp 3bp 0bp 3bp},clip]{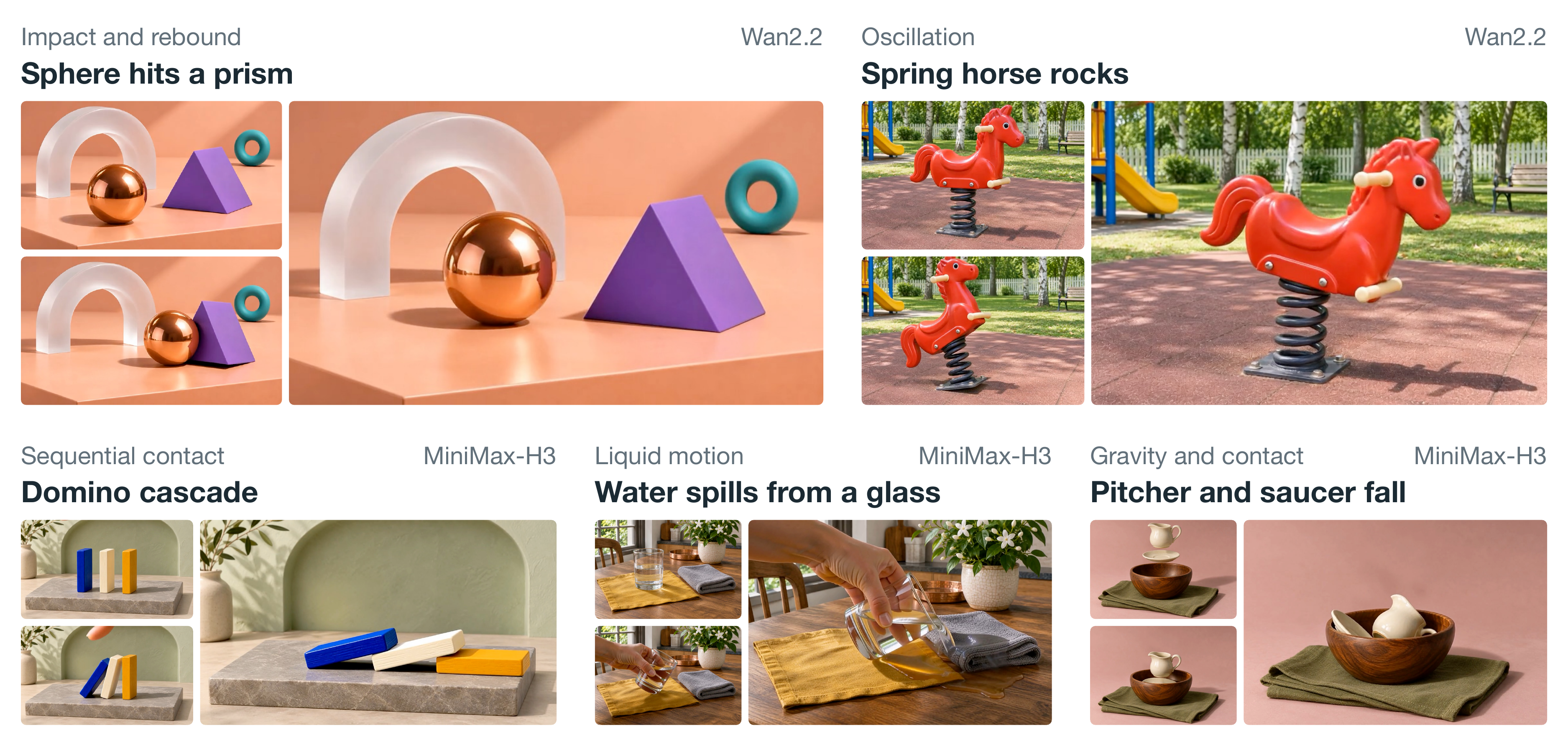}
  \caption{\textbf{Videos generated with FantasyEvo.} For each task, the agent writes the prompt with the harness learned for the target VDM (top right of each panel). Each panel shows frames from the generated video, with the last frame enlarged, across five types of physical events: impact and rebound, oscillation, sequential contact, liquid motion, and gravity and contact.}

\label{fig:model-teaser}
\end{minipage}\par

\section{Introduction}
\label{sec:intro}

\begin{figure*}[t]
\centering
\includegraphics[width=\textwidth]{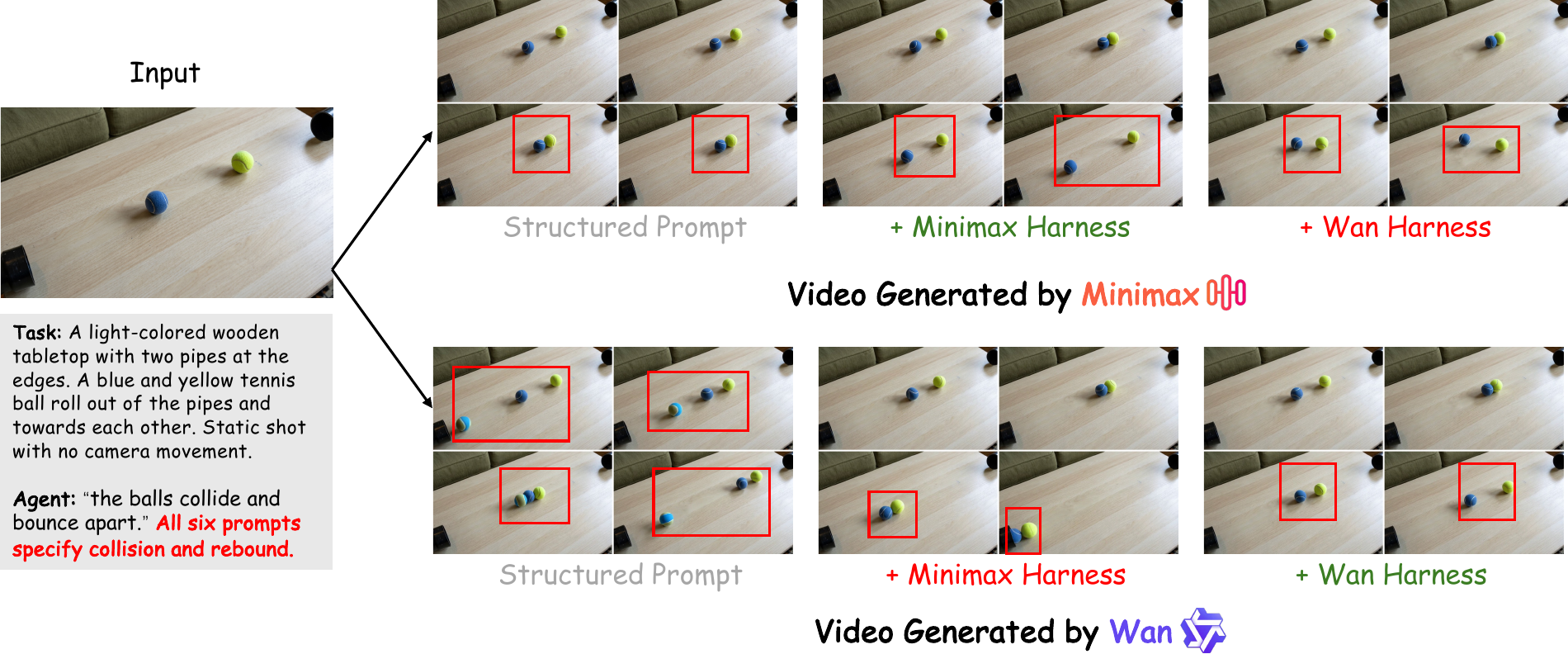}
\caption{\textbf{Knowing is not telling.} The same agent writes a separate prompt for each combination of three harnesses and two video models, yielding six prompts. Each model renders its three prompts once: MiniMax-H3 (top) and Wan2.2 (bottom). The generic structured harness fails on both models. Each learned harness succeeds on the model it was learned for (green) but fails on the other (red). Red boxes mark where the two balls should meet.}
\label{fig:teaser}
\end{figure*}


Video diffusion models (VDMs) have improved quickly in recent years~\cite{seedance,brooks2024video,veo3,kling,kong2024hunyuanvideo,wan2025wan,agarwal2026cosmos,minimaxh3}. Their videos now look more and more real, with sharp detail, natural lighting and smooth camera motion.
However, they still do poorly on tasks that need physical or logical reasoning, such as predicting where a falling object lands or how a hidden state keeps changing while it is out of view~\cite{motamed2026generative,ma2026out,bansal2025videophy,bansal2026videophy,meng2024towards}.
The problem lies not in how a VDM draws each frame, but in how it decides what should happen from one frame to the next. Getting this right means working out causes and effects in the scene, which is a form of reasoning.
Pairing a VDM with a reasoning agent seems a natural fix: the agent supplies the physics, and the VDM renders it. 

Existing methods put this idea into practice in several ways. These methods fall into two paradigms.
The first is \emph{describe-then-render}: the agent rewrites the task into a richer description, adding the events and physical details that the VDM would otherwise miss, and may further revise it using feedback on the rendered video or pass part of it as extra conditioning such as an end frame~\cite{wan2025wan,agarwal2026cosmos,zhou2026tide,xue2025phyt2v,miao2026video,lian2024llm,lin2023videodirectorgpt}. The second is \emph{simulate-then-render}, in the spirit of code world models~\cite{tang2024worldcoder,dainese2024generating,zhang2026code2worlds,chen2026code,li2026recursive}: the agent writes executable code, such as a physical simulation or a calculation of physical quantities, and its result then guides rendering~\cite{lv2024gpt4motion,liu2024physgen,feng2026newton}.
In both paradigms, the agent supplies what VDMs lack, an explicit account of causes, order and physical quantities, and the VDM supplies realistic pixels. Yet both take for granted that the agent's knowledge reaches the VDM, and this step is often where things go wrong.

Fig.~\ref{fig:teaser} shows an example. Two tennis balls roll toward each other. The agent knows they should collide and bounce apart, and every prompt says so. Still, with a generic structured prompt, neither MiniMax-H3 nor Wan2.2 shows the collision. The prompt learned for MiniMax-H3 works on MiniMax-H3 but not on Wan2.2. The prompt learned for Wan2.2 works only on Wan2.2. Knowing is not telling, and how to tell depends on the VDM.
We observe the same gap between what the prompt states and what the video shows on StEvo with MiniMax-H3. On the 102 tasks where both prompts fully and correctly state the expected hidden-state change, our harness raises the pass rate by 24.5 percentage points. Across all 221 tasks, the pass rate rises by 15.8 percentage points, with 25 of the 35 net additional passes coming from this subset (Sec.~\ref{sec:analysis}).

Therefore, we propose FantasyEvo, which addresses this with a learned \emph{harness}: a set of readable rules that tells the agent which information to include, how to organize it, and how to phrase it for one specific VDM.
The harness targets the step that prior work leaves fixed. The agent and the VDM stay unchanged, and only the way the agent's knowledge is written into the prompt is learned.
Because the right way to tell depends on the receiver and is not written down anywhere, we learn a separate harness for each VDM from that model's own feedback. Starting from a generic structured rewrite, an evolver reads which success criteria failed on development tasks, that is, which pieces of information did not show up in the video, and proposes one revision of the rules at a time. A revision is kept only if it improves results on held-out tasks. Unlike feedback loops that start over on every task, this learning happens once per VDM and offline. The rules are written to hold for any task, so the harness is then frozen and reused.
At test time, each new task costs one agent call and one VDM call, the same as a plain rewrite. It needs no extra inputs or simulator and works on any scene the agent can describe. Fig.~\ref{fig:model-teaser} shows some examples on MiniMax-H3 and Wan-2.2 using our FantasyEvo, where the videos follow the expected physical outcome.

We evaluate on three VDMs (MiniMax-H3~\cite{minimaxh3}, Wan2.2~\cite{wan2025wan}, Cosmos3-Nano~\cite{agarwal2026cosmos}) and two benchmarks (Physics-IQ~\cite{physicsiq}, StEvo~\cite{ma2026out}). Our harness achieves the best overall score in all six settings, surpassing the strongest physics-aware method~\cite{zhou2026tide,xue2025phyt2v,feng2026newton} by 3.4 and 13.3 points on average. Harnesses are also receiver-specific: each VDM performs best with its own, and transferred harnesses lose 3.9 points on Physics-IQ.

In summary, our contributions are as follows:
    \begin{itemize}
    \item We show that in agent-assisted video generation, much of what the agent knows is lost on its way to the VDM, and that how it should be told depends on the receiving VDM.
    \item We propose a VDM-specific harness, learned once from the VDM's own feedback and then frozen, which tells the agent how to write its knowledge for that model at the cost of one agent call and one VDM call per task.
    \item On three VDMs and two benchmarks, the harness improves over the raw prompt by 8.5 and 24.7 points on Physics-IQ and StEvo, and surpasses the strongest physics-aware method by 3.4 and 13.3 points.
\end{itemize}

\section{Related Work}
\label{sec:formatting}

\paragraph{Physics in video generation.} Video diffusion models produce realistic frames but often violate basic physics, as shown by a growing set of benchmarks~\cite{bansal2025videophy,bansal2026videophy,meng2024towards,motamed2026generative,kang2024howfar,worldmodelbench,vbench2}. StEvo~\cite{ma2026out} further shows that a state stops evolving once it is hidden from view. To close this gap, prior work couples generation with physical simulation~\cite{physdreamer,motioncraft}, trains on physics-aware data or objectives~\cite{wisa,physmaster,videorepa}, or adds explicit physical controls such as forces~\cite{forceprompting}. All of these change the video model or its inputs beyond text, while we keep the model frozen and change only how the task is told to it.

\paragraph{LLM agents for video generation.} LLMs are widely used to plan layouts, scene graphs or shot-level descriptions before generation~\cite{lin2023videodirectorgpt,lian2024llm,freebloom,vlogger,dreamfactory}. For physical plausibility, existing agents follow two paradigms. \emph{Describe-then-render} methods rewrite the task into a richer description: PhyT2V~\cite{xue2025phyt2v} and DiffPhy~\cite{diffphy} refine prompts with feedback on rendered videos, StateAgent~\cite{miao2026video} renders a predicted end state as the last frame, and TIDE~\cite{zhou2026tide} distills a rewriting skill that it applies unchanged across models. \emph{Simulate-then-render} methods, in the spirit of code world models~\cite{tang2024worldcoder,dainese2024generating}, compute physics with code or a simulator and use the result to guide rendering~\cite{lv2024gpt4motion,liu2024physgen,feng2026newton}. Each paradigm has its own cost: feedback loops render each task several times and carry nothing over to the next task, and simulators are hard to apply to liquids, soft objects or long chains of events. More importantly, both focus on what the agent knows and leave the final step of telling it to the VDM fixed, as a generic rewrite or a template shared by all models. We show that this step loses much of what the agent knows, and learn a model-specific harness for it at no extra test-time cost.

\paragraph{Learning prompts from feedback.} A broad family of methods optimizes natural-language prompts with model feedback, through search~\cite{ape,protegi,promptagent}, LLMs as optimizers~\cite{opro}, evolution~\cite{evoprompt,promptbreeder,gepa}, textual gradients~\cite{textgrad}, or program compilation~\cite{dspy,mipro}, and agents can also accumulate reusable lessons, skills or playbooks from experience~\cite{reflexion,expel,voyager,awm,dynamiccheatsheet,ace}. For visual generation, prompt rewriters are trained with aesthetic or alignment rewards~\cite{promptist,promptavideo,vpo} or retrieve modifiers that match the training distribution~\cite{rapo}. Our learning procedure belongs to this family, and we do not claim a new optimizer. 
We keep the original task inputs fixed and use feedback on individual rubric checks to learn what to include in a prompt and how to express it for each video model.

\section{Method}
\label{sec:method}

\subsection{Problem Formulation and Overview}
\label{sec:overview}



\paragraph{Problem formulation.}
We consider image-conditioned video generation. A task $x=(I_0,c)$ consists of a first frame $I_0$ and a description $c$ of the requested action. Given a text prompt $p$, a VDM $G_v$ generates a video $y=G_v(I_0,p)$, and a score $S(y;x)$ measures how well $y$ shows the action and what physically follows. Instead of using $p=c$ directly, an agent $\mathcal{A}$ reads the task and writes the prompt $p=\mathcal{A}(x;H)$, where the harness $H$ includes everything that shapes the agent's output, such as its instructions, examples, and output format. With $\mathcal{A}$ and $G_v$ fixed, we seek for each VDM the harness that maximizes the expected score over tasks $x\sim\mathcal{D}$ and seeds:
\begin{equation}
H_v^\star = \arg\max_{H}\; \mathbb{E}_{x\sim\mathcal{D},\, s}\big[S\big(G_v(I_0, \mathcal{A}(x;H); s);\, x\big)\big].
\label{eq:objective}
\end{equation}

\begin{figure*}[t]
  \centering
\includegraphics[width=\linewidth]{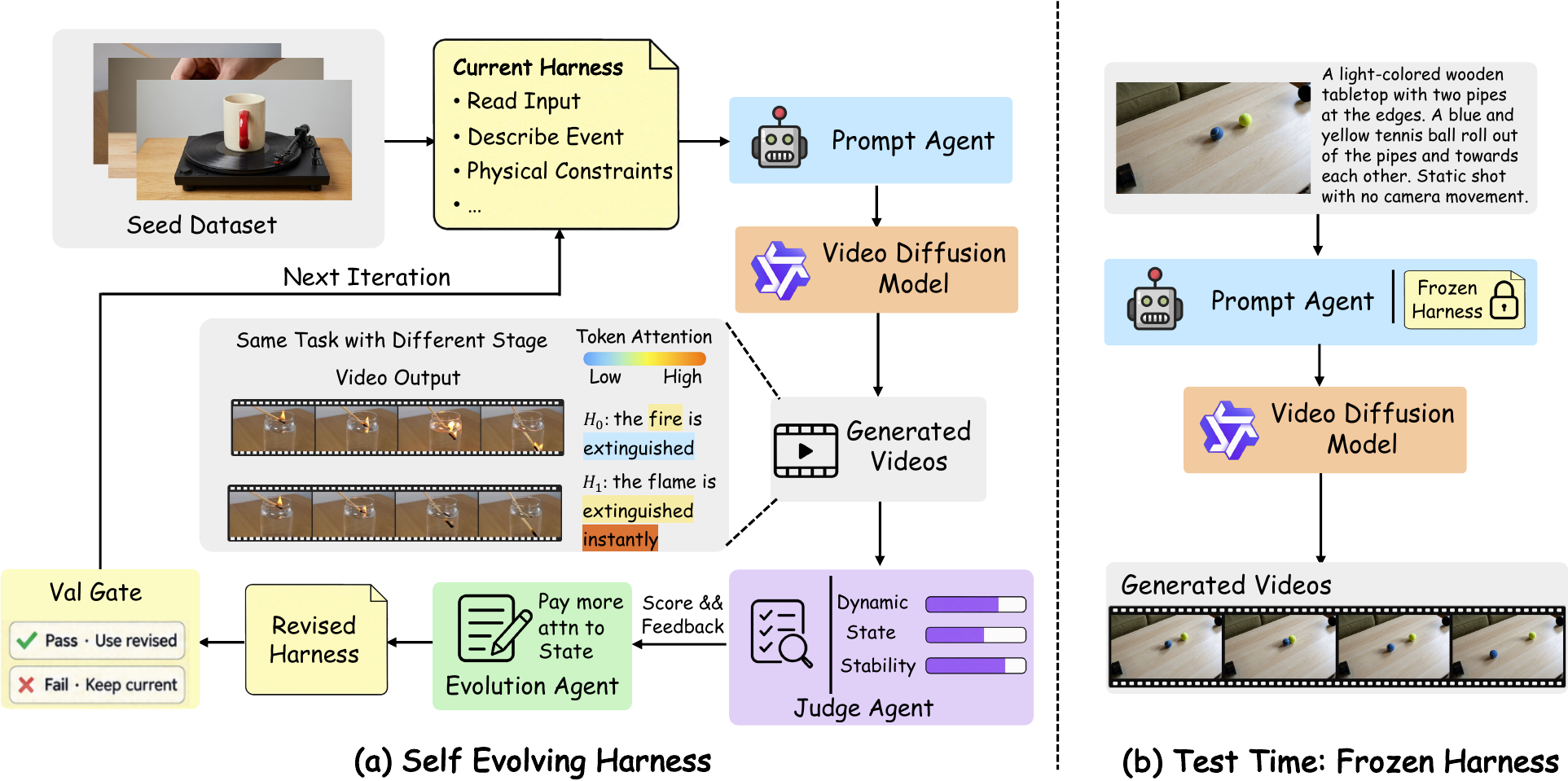}
\caption{\textbf{Overview of our method.} (a) \emph{Self-evolving harness.} The prompt agent writes prompts for the seed tasks under the current harness, the VDM renders them, and the judge agent scores each video with textual feedback. The evolution agent then proposes a revised harness, which is kept only if it passes the validation gate. Inset: after one revision, the prompt states that the flame goes out ``instantly'', and the VDM's attention on this word rises. (b) \emph{Test time.} With the harness frozen, each task takes one prompt and one video.}
  \label{fig:method}
\end{figure*}

\paragraph{Overview.}
Existing pipelines fix $H$ by hand, for example by using the task description as is~\cite{physicsiq,ma2026out} or applying a generic rewrite instruction~\cite{kong2024hunyuanvideo,wan2025wan,agarwal2026cosmos}. Such prompts are more detailed, but their key information does not necessarily appear in the video: whether a VDM renders it depends on how it is written, and a phrasing that one VDM follows may be ignored by another. We therefore learn a separate harness $H_v$ for each VDM, a natural-language document of rules that tells the prompt agent which information to include and how to organize, phrase, and ground it in the first frame (Sec.~\ref{sec:harness}).

As shown in Fig.~\ref{fig:method}(a), the harness is learned offline on a small seed dataset covering common physical phenomena (Sec.~\ref{sec:evolve}). In each iteration, the prompt agent writes prompts under the current harness, the VDM renders them, and a judge agent scores each video check by check against the task's rubric. An evolution agent reads the failed checks, diagnoses where the information was lost, and proposes a revised harness, which replaces the current one only if it passes a validation gate, i.e., its paired gain on validation tasks exceeds the sampling noise. At test time (Fig.~\ref{fig:method}(b)), the harness is frozen, and each task costs one prompt-agent call and one VDM call, the same as a zero-shot rewrite.

\subsection{What to Learn: The Harness}
\label{sec:harness}


Generating the intended video using video diffusion models requires two kinds of knowledge: what should happen in the scene, and how the target VDM turns text into video. The prompt agent, a multimodal LLM, can work out the first from the frame and the task. It lacks the second, since it has never seen which information a particular VDM renders or ignores, which layouts and wordings it follows, or how it reconciles the text with the first frame. 
The two kinds of knowledge also differ in scope: what should happen changes with every task, whereas how a VDM reads a prompt stays largely the same across tasks. We therefore learn the second kind of knowledge for each VDM and write it down as a \emph{harness}, a plain-text document of rules that the prompt agent follows, which is the $H_v^*$ in Eq.~\eqref{eq:objective}.

The harness covers four aspects of writing a prompt: what information to include, how to organize it, how to phrase it, and how to keep it consistent with the first frame. Since one harness serves all tasks of a VDM, its rules must be general and may not refer to a specific task. All VDMs start from the same hand-written harness $H^{(0)}$ with generic rules for each aspect, and learning may add, revise, or remove rules of any kind, as described next.

\subsection{How to Learn: Self-Evolving Harness}
\label{sec:evolve}


The generic rules in $H^{(0)}$ do not reflect how a particular VDM reads a prompt. The prompt agent lacks this knowledge (Sec.~\ref{sec:harness}), and official prompt guides specify the input format but not which information the VDM tends to drop, so it must be discovered from the VDM's own outputs. We do so with a self-evolving loop, shown in Fig.~\ref{fig:method}(a). In each iteration, the prompt agent writes a prompt for each seed task under the current harness $H_v^{(t)}$, initialized as $H^{(0)}$; the VDM renders the prompts into videos; and the judge agent scores each video and returns feedback. The evolution agent reads this feedback and proposes a revised harness $H'$. A validation gate then either accepts $H'$ as $H_v^{(t+1)}$ or rejects it and keeps $H_v^{(t)}$ for the next iteration. We describe the seed dataset, judge agent, evolution agent, and validation gate below.

\paragraph{Seed dataset.}
The evolution loop needs tasks on which we can tell which pieces of information a video fails to show. We build 52 seed tasks, each with a first frame $I_0$, a task description $c$, and a rubric $R(x)=\{r_k\}$ of $K$ observable checks. The tasks cover both event ordering and physical outcomes, and share no scene or object with any evaluation set. 
For each task, a design agent writes the rubric together with the description, and an image model~\cite{zhao2026qwen} then renders $I_0$ from that description; a review agent~\cite{qwen38} either accepts the frame or sends it back for revision, which keeps $I_0$ consistent with the rubric. 
We split the tasks into a development set $\mathcal{D}_\text{dev}$ of 36 tasks, whose failures guide the revisions, and a validation set $\mathcal{D}_\text{val}$ of 16 tasks, which only decides whether a revision is kept.

\paragraph{Judge Agent.}
Each rubric check $r_k$ is an observable statement for one piece of information the video should convey, so a failed check shows exactly which information the video did not render. The judge agent scores each check on frames sampled across the video, giving a grade $q_k \in \{0, 0.25, 0.5, 0.75, 1\}$ and a short note $e_k$ on what the video shows (e.g., ``the ball stops short of the bowl''). The evolution agent later uses these notes to diagnose failures. 
Each check is judged three times and the median is kept. Checks are grouped into action checks $\mathcal{K}_a$ (the requested action happens), outcome checks $\mathcal{K}_o$ (its physical result follows), and integrity checks $\mathcal{K}_c$ (the rest of the scene stays intact), and the video score is
\begin{equation}
S(y;x) = G(\mathcal{K}_a)\cdot G(\mathcal{K}_o)\cdot M(\mathcal{K}_c),
\label{eq:score}
\end{equation}
where $G(\cdot)$ and $M(\cdot)$ are the geometric and arithmetic means of $q_k$ over a group. A single fully failed action or outcome check therefore sets the score to zero, while integrity checks only lower it.


\paragraph{Evolution Agent.}
A failed check shows that information was lost but not where: the prompt may have omitted it, stated it wrongly, or stated it correctly while the VDM failed to render it. The evolution agent therefore diagnoses before it revises. For each task in $\mathcal{D}_\text{dev}$, it reads the prompt, the scores by check type, and every check scoring below $0.75$ with its note, and decides which case applies. Some failures cannot be fixed through the harness, such as a detail that the sampled frames cannot show or an effect that the VDM cannot render under any prompt. The agent records these and leaves them out. For the rest, it proposes a single revision that adds, replaces, or merges rules, or changes the layout while keeping the information. Rejected revisions stay in its history so that it does not repeat them. 

Fig.~\ref{fig:method}(a) shows the effect of one such revision on the same task. Under $H^{(0)}$, the prompt says that the fire ``is extinguished'', and the VDM pays little attention to the word carrying the outcome. After the revision, $H^{(1)}$ states that the flame is extinguished ``instantly'', and the attention on this word rises. The revision thus changes how the VDM reads the information, not only what the prompt contains.


\paragraph{Validation Gate.}
Not every revision proposed by the evolution agent helps, and a higher validation score alone does not show that it does. 
The gate therefore compares $H'$ with $H_v^{(t)}$ on $n$ matched pairs, each consisting of a validation task and a VDM seed. For each pair, the prompt agent writes a prompt under each harness, the VDM renders both prompts with the pair's seed, and the judge agent scores the two videos. $H'$ is accepted only if the score gain averaged over the $n$ pairs reaches a threshold $\delta$. Since the two videos in a pair share the task and the seed, task difficulty and seed noise cancel in the difference, so the gain reflects the change in the rules rather than luck.
\section{Experiments}
\label{sec:exp}

We organize the experiments around four questions.

\begin{itemize}[leftmargin=*,itemsep=1pt,topsep=2pt]
    \item \textbf{RQ1 (Main results).} Does a learned harness deliver the agent's knowledge to the VDM better than other prompting methods, including those that spend more renders per task (Sec.~\ref{sec:main})?
    \item \textbf{RQ2 (Receiver specificity).} Is what a harness learns specific to the VDM it was learned on, or does it also work on other VDMs (Sec.~\ref{sec:swap})?
    \item \textbf{RQ3 (Information loss).} Does the gain come from the agent stating the expected change more often, or from the VDM rendering it more often once it is stated (Sec.~\ref{sec:analysis})?
    \item \textbf{RQ4 (Ablation).} How does the harness improve over evolution rounds, and how sensitive is learning to the models used as the prompt, judge, and evolution agents (Sec.~\ref{sec:ablation})?
\end{itemize}


\subsection{Setup}
\label{sec:setup_exp}

\paragraph{Video models and agent.} We use three image-to-video models: MiniMax-H3~\cite{minimaxh3}, Wan2.2-I2V-A14B~\cite{wan2025wan}, and Cosmos3-Nano~\cite{agarwal2026cosmos}. All videos are 5 seconds long and are rendered at each model's native setting, 24\,fps for MiniMax-H3 and Cosmos3-Nano and 16\,fps for Wan2.2-I2V-A14B. The same agent, \texttt{qwen3.8-max}, takes the task text, the first frame of the video, and a model-specific harness as input, and returns a prompt that is passed to the target VDM.

\paragraph{Baselines.} We compare against two groups of baselines (Table~\ref{tab:main}). \textbf{Generic prompting}: \emph{Raw} passes the task text to the VDM unchanged, \emph{Official} uses the VDM's own prompt-rewriting instructions, and \emph{Structured} uses our hand-written harness $H^{(0)}$, the starting point of learning. \textbf{Physics-aware prompting}: \emph{TIDE}~\cite{zhou2026tide} applies a frozen, published rewriting skill, to which we add a first-frame description since it was built for text-to-video; \emph{PhyT2V}~\cite{xue2025phyt2v} refines the prompt with feedback on rendered videos (3 renders per task); and \emph{OpenNewton}~\cite{feng2026newton} plans with a calculator and re-renders under VLM verification (1 to 3 renders). All other methods, like ours, use one render per task

\paragraph{Benchmarks and metrics.}
We evaluate on Physics-IQ~\cite{physicsiq} (198 tasks) and StEvo~\cite{ma2026out} (221 tasks). Both benchmarks are held out from harness learning and version selection. Physics-IQ tests whether a generated video correctly continues a real recording. Each generated continuation is compared with the real footage that follows the first frame. We report the official Physics-IQ score, which combines spatial, spatiotemporal, and weighted spatial IoU and subtracts an MSE term. StEvo tests whether a state continues to evolve while it is hidden by occlusion, a camera look-away, or dimmed lighting. 
A video passes if the hidden state evolves correctly without physical inaccuracies. The judge assesses each of these two criteria three times and decides each by majority vote; we report the percentage of videos that pass, using the same protocol for all methods.

\paragraph{Harness learning.} Harnesses are learned on the seed tasks of Sec.~\ref{sec:evolve}, with $\mathcal{D}_\text{dev}$ guiding the revisions and $\mathcal{D}_\text{val}$ deciding acceptance through the validation gate. The prompt agent and the judge agent are both qwen3.8-max~\cite{qwen38}, and the evolution agent is DeepSeek-V4-Pro~\cite{xu2026deepseek}. The frozen harness is reached after 10, 26, and 29 attempts for MiniMax-H3, Wan2.2, and Cosmos3-Nano, and each accepted version is rendered once more on $\mathcal{D}_\text{val}$. The harness evaluated on the benchmarks is the last version accepted on $\mathcal{D}_\text{val}$, frozen before any benchmark video is generated.


\subsection{Main Results}
\label{sec:main}

\begin{table*}[t]
\centering
\definecolor{kgpos}{RGB}{0,130,50}
\definecolor{kgneg}{RGB}{200,20,20}
\newcommand{\dl}[1]{\,{\scriptsize\textcolor{kgpos}{$+$#1}}}
\newcommand{\ds}[1]{\,{\scriptsize\textcolor{kgneg}{$-$#1}}}
\caption{Main results (higher is better). Small numbers give the change from Raw (\textcolor{kgpos}{green}: increase, \textcolor{kgneg}{red}: decrease). Best in each row in \textbf{bold}, second best \underline{underlined}}
\label{tab:main}
\small
\setlength{\tabcolsep}{3pt}
\renewcommand{\arraystretch}{1.0}
\begin{tabular*}{\textwidth}{@{\extracolsep{\fill}}ll ccc ccc c}
\toprule
 & & \multicolumn{3}{c}{Generic prompting} & \multicolumn{3}{c}{Physics-aware prompting} & \\
\cmidrule(lr){3-5}\cmidrule(lr){6-8}
Benchmark & VDM & Raw & Official & Structured & TIDE & PhyT2V & OpenNewton & \textbf{Ours} \\
 & & & & & \cite{zhou2026tide} & \cite{xue2025phyt2v} & \cite{feng2026newton} & \\
\midrule
\multirow{3}{*}{Physics-IQ}
 & MiniMax-H3   & 39.5 & 43.3\dl{3.8} & \underline{44.9}\dl{5.4} & 44.6\dl{5.1} & 41.4\dl{1.9} & 42.0\dl{2.5} & \textbf{48.3}\dl{8.8} \\
 & Wan2.2       & 27.6 & 32.7\dl{5.1} & \underline{34.4}\dl{6.8} & 33.7\dl{6.1} & 33.4\dl{5.8} & 32.3\dl{4.7} & \textbf{37.3}\dl{9.7} \\
 & Cosmos3-Nano & 30.1 & \underline{35.6}\dl{5.5} & 34.0\dl{3.9} & 34.3\dl{4.2} & 32.2\dl{2.1} & 34.4\dl{4.3} & \textbf{37.2}\dl{7.1} \\
\midrule
\multirow{3}{*}{StEvo}
 & MiniMax-H3   & 30.3 & \underline{56.1}\dl{25.8} & 51.6\dl{21.3} & 33.5\dl{3.2} & 46.2\dl{15.9} & 52.9\dl{22.6} & \textbf{67.4}\dl{37.1} \\
 & Wan2.2       & 46.2 & 42.5\ds{3.7} & 45.2\ds{1.0} & 42.1\ds{4.1} & 43.4\ds{2.8} & \underline{49.8}\dl{3.6} & \textbf{59.7}\dl{13.5} \\
 & Cosmos3-Nano & 23.5 & \underline{40.7}\dl{17.2} & 38.9\dl{15.4} & 24.4\dl{0.9} & 27.6\dl{4.1} & 31.7\dl{8.2} & \textbf{47.1}\dl{23.6} \\
\bottomrule
\end{tabular*}
\end{table*}

\textbf{Quantitative Results.} Table~\ref{tab:main} reports results on Physics-IQ and StEvo. Our harness obtains the best overall score on all three VDMs and both benchmarks, while the strongest baseline varies with the VDM and the benchmark (Structured, Official, or OpenNewton). Compared with the best-performing baseline among all methods in each setting, FantasyEvo improves Physics-IQ by 3.4, 2.9, and 1.6 points and StEvo by 11.3, 9.9, and 6.4 points on MiniMax-H3, Wan2.2, and Cosmos3-Nano, respectively. Rewriting the prompt does not help by itself: on Wan2.2, four of the five baselines score below Raw on StEvo, whereas our harness exceeds Raw by 13.5 points. A harness learned for each VDM improves generation across models and benchmarks at the same test-time cost as a single rewrite. It also beats PhyT2V and OpenNewton, which spend up to three renders per task, in all six settings, so the gain comes from how the prompt is written rather than from extra test-time compute.

\begin{figure*}[t]
\centering
\includegraphics[width=\textwidth]{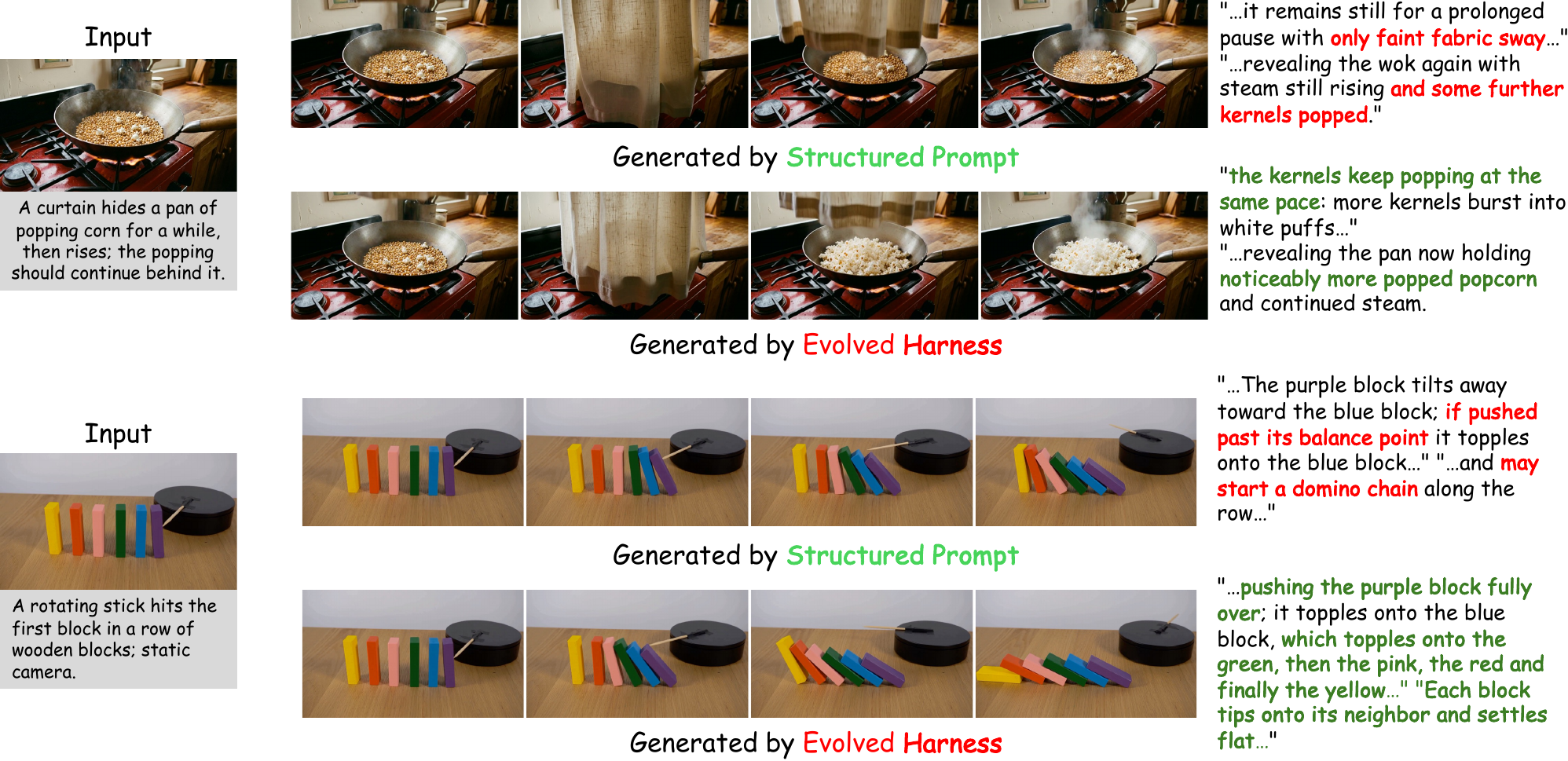}
\caption{\textbf{Qualitative comparison on MiniMax-H3}. Each pair uses the same first frame. \textbf{Right}: excerpts from the prompts that the agent writes, with the parts that differ highlighted. \textbf{Top}: a curtain hides a pan of popping corn. \textbf{Bottom}: a rotating stick hits the first block.}
\label{fig:quali}
\end{figure*}

\begin{figure*}[t]
\centering
\includegraphics[width=\textwidth]{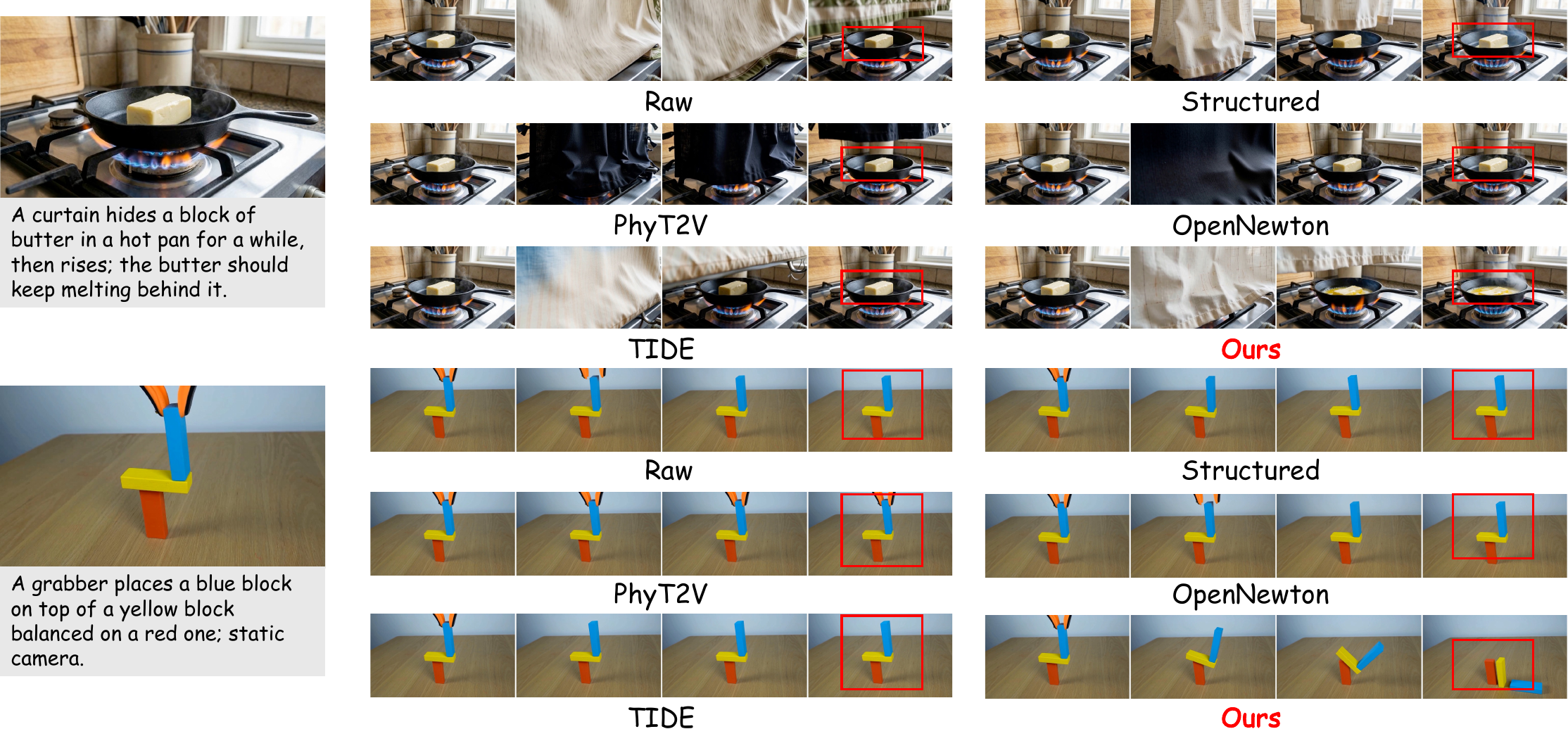}
\caption{\textbf{Qualitative comparison on MiniMax-H3 with baseline.} Left: the first frame and the task. Right: four frames from each method. Red boxes mark where the expected change should appear. Top: a block of butter keeps melting while a curtain hides it. Bottom: the block stack should topple after the grabber places the blue block.}
\label{fig:quali1}
\end{figure*}

\textbf{Qualitative Results.} Fig.~\ref{fig:quali} contrasts the structured harness with our evolved harness on MiniMax-H3, with the prompt excerpts that differ. In the popcorn example, a curtain hides the pan. The structured prompt spends its words on the curtain (``only faint fabric sway'') and hedges on the hidden change (``some further kernels popped''), and the pan looks almost unchanged when the curtain rises. The evolved prompt states the change directly (``the kernels keep popping at the same pace''), and the revealed pan is full of popcorn. In the block-row example, the structured video ends with the blocks leaning on each other, while the evolved video ends with every block fallen flat. Fig.~\ref{fig:quali1} adds the other baselines~\cite{zhou2026tide,feng2026newton,xue2025phyt2v}. In the butter example, every baseline shows the same solid block when the curtain rises, so the melting stops once the butter is out of view, and only ours shows it melted into a puddle. In the block example, every baseline keeps the stack standing after the grabber places the blue block, while in ours the stack tips over and the blue block lands on the table. In these examples the baselines leave the scene in its initial state, whereas the harness makes the expected change visible.

\subsection{Are Harnesses Receiver-Specific?}
\label{sec:swap}



  \begin{table*}[t]
  \definecolor{kgpos}{RGB}{0,130,50}
  \definecolor{kgneg}{RGB}{200,20,20}
  \newcommand{\kgcell}[3]{%
    \begin{tabular}[c]{@{}c@{}}
      #1#3\\[-1pt]
      {\scriptsize\textcolor{black!65}{(#2 / 221)}}
    \end{tabular}%
  }
  \newcommand{\kgup}[1]{\,{\scriptsize\bfseries\textcolor{kgpos}{$+$#1}}}
  \newcommand{\kgdown}[1]{\,{\scriptsize\bfseries\textcolor{kgneg}{$-$#1}}}
  \setlength{\tabcolsep}{3pt}
  \begin{minipage}[t]{0.48\textwidth}
  \centering
  \caption{Harness swap on Physics-IQ. Each harness (row) is used to prompt each VDM, with the same prompt agent and tasks. Diagonal entries use the harness learned on that VDM. Best result in each column in \textbf{bold}.}
  \label{tab:swap}
  \footnotesize
  \renewcommand{\arraystretch}{1.1}
  \begin{tabular*}{\linewidth}{@{\extracolsep{\fill}}lccc@{}}
  \toprule
  Harness & MiniMax & Wan & Cosmos \\
  \midrule
  $H^{(0)}$ (Structured) & 44.9 & 34.4 & 34.0 \\
  \midrule
  $H^\star_{\text{MiniMax}}$ & \textbf{48.3} & 35.3 & 34.5 \\
  $H^\star_{\text{Wan}}$ & 47.1 & \textbf{37.3} & 33.0 \\
  $H^\star_{\text{Cosmos}}$ & 42.3 & 30.0 & \textbf{37.2} \\
  \bottomrule
  \end{tabular*}
  \end{minipage}\hfill
  \begin{minipage}[t]{0.48\textwidth}
  \centering
  \caption{Prompt coverage and video outcomes on StEvo with MiniMax-H3.
  \emph{Told} means the prompt fully and correctly states the expected
  hidden-state change. Colored values show percentage-point changes
  relative to Structured.}
  \label{tab:funnel}
  \footnotesize
  \renewcommand{\arraystretch}{0.95}
  \begin{tabular}{@{}lcccc@{}}
  \toprule
  & \multicolumn{2}{c}{\textbf{Structured}}
  & \multicolumn{2}{c}{\textbf{Ours}} \\
  \cmidrule(lr){2-3}\cmidrule(lr){4-5}
  & Wrong & Right & Wrong & Right \\
  \midrule
  \textbf{Not told}
    & \kgcell{26.70}{59}{}
    & \kgcell{8.14}{18}{}
    & \kgcell{25.79}{57}{\kgdown{0.90}}
    & \kgcell{6.33}{14}{\kgdown{1.81}} \\
  \addlinespace[1pt]
  \textbf{Told}
    & \kgcell{21.72}{48}{}
    & \kgcell{43.44}{96}{}
    & \kgcell{6.79}{15}{\kgdown{14.93}}
    & \kgcell{61.09}{135}{\kgup{17.65}} \\
  \bottomrule
  \end{tabular}
  \end{minipage}
  \end{table*}
If the harness only taught the agent to write better prompts in general, a harness learned on one VDM should help the others as well. We test this by keeping the agent and the tasks fixed and prompting every VDM with each learned harness (Table~\ref{tab:swap}). In every column, the harness learned on that VDM is best, improving over $H^{(0)}$ by 3.4, 2.9, and 3.2 points on MiniMax-H3, Wan2.2, and Cosmos3-Nano. A harness from another VDM loses 3.9 points on average relative to the VDM's own, and in three of the six swaps it falls below $H^{(0)}$. Fig.~\ref{fig:teaser} shows the same effect on one task. Every prompt describes the collision and rebound, yet the structured prompt fails on both VDMs, and each VDM shows the rebound only under the harness learned on it. A way of writing that helps one model can thus be neutral or harmful for another.

\subsection{Where Is Information Lost?}
\label{sec:analysis}

\textbf{Settings.}
We ask whether the gain on StEvo comes from what the agent writes into the prompt or from how the VDM responds to it. We use MiniMax-H3 and the videos behind for two harnesses: Structured, the harness from which our evolution starts, and Ours. We record two labels per task and harness. \emph{Told}: a prompt counts as told if it correctly states what should happen, as judged by an LLM against the benchmark's ground truth. \emph{Right}: the video passes StEvo's overall criterion, with correct hidden-state evolution and no physical inaccuracies.

\begin{figure*}[t]
\centering
\includegraphics[width=\textwidth]{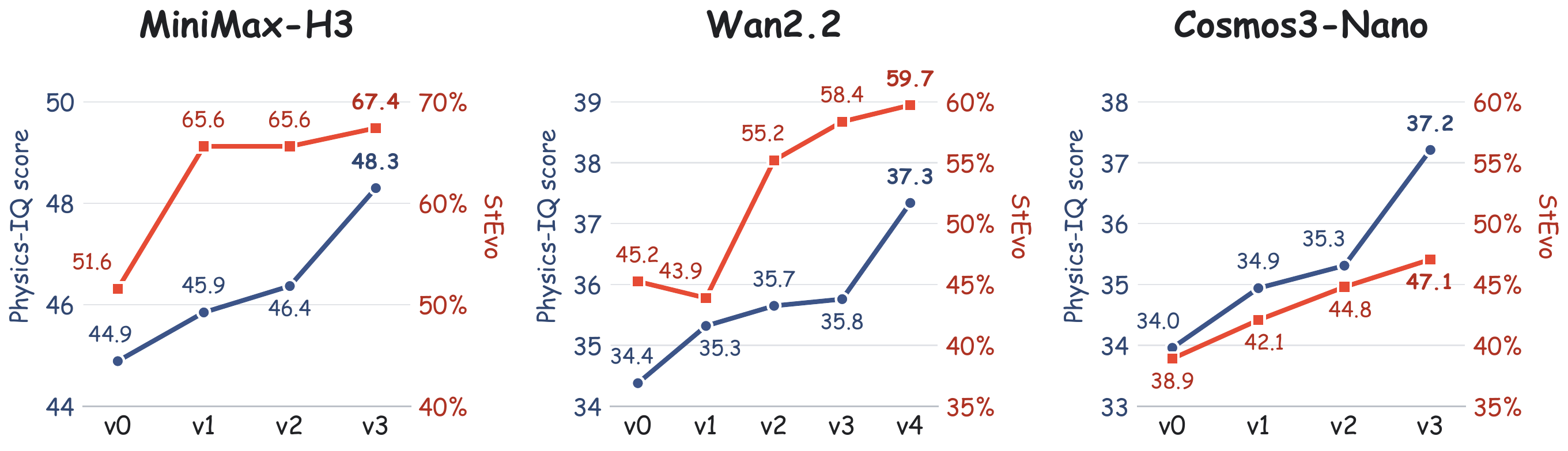}
\caption{\textbf{Performance across evolution rounds.} Physics-IQ score (blue, left axis) and StEvo (red, right axis) of the harness after each round of evolution on MiniMax-H3, Wan2.2, and Cosmos3-Nano. v0 is the generic structured rewrite that evolution starts from.}
\label{fig:evolution_rounds}
\end{figure*}

\textbf{Results.} Table~\ref{tab:funnel} shows that our harness raises the number of told tasks only modestly, from 144 (65.2\%) to 150 (67.9\%), but told prompts are successfully rendered far more often: their failure rate falls from 33.3\% (48/144) to 10.0\% (15/150). We isolate the failure mode of Fig.~\ref{fig:teaser} on the 102 tasks whose prompts are told under both harnesses. Across these 102 tasks, 26 went from failing under Structured to passing under our harness, and only one went from passing to failing. This yields a net gain of 25 passes, or 24.5 percentage points.
Because both prompts state the same change on these tasks, the difference lies mainly in how the change is written rather than whether it is stated. The gain thus comes from delivering information that the prompt already contains in a form the VDM renders.

\subsection{Ablations}
\label{sec:ablation}


\textbf{Ablation on harness versions.} Fig.~\ref{fig:evolution_rounds} reports the harness after each accepted version, starting from the structured harness (v0). For all three VDMs, the final version scores highest on both benchmarks, improving Physics-IQ by 3.4, 2.9, and 3.2 points and StEvo by 15.8, 14.5, and 8.2 points on MiniMax-H3, Wan2.2, and Cosmos3-Nano, respectively. The two benchmarks improve differently. On StEvo, most of the gain on MiniMax-H3 and Wan2.2 comes from a single revision (+14.0 at v1 and +11.3 at v2, respectively), and each subsequent revision adds at most 3.2 points. On Physics-IQ, the gains are smaller and spread across versions, with the largest gain coming from the final revision for each model (+1.9, +1.5, and +1.9 points, respectively). This suggests that StEvo failures are dominated by a few shared issues that one revision can fix, whereas Physics-IQ covers more varied events whose fixes accumulate.

\begin{table}[t]
\centering
\caption{\textbf{Ablation on agent type.} The bottom rows replace the model of one agent during harness evolution (underlined) and evaluate the resulting harness on MiniMax-H3. Both benchmarks report the overall score. Max and 27B denote Qwen3.8-Max and Qwen3.8-27B; DS-V4 denotes DeepSeek-V4-Pro.}
\label{tab:ablation_agent_type}
\small
\setlength{\tabcolsep}{3.5pt}
\begin{tabular}{lccccc}
\toprule
 & \multicolumn{3}{c}{Agent model} & \multicolumn{2}{c}{Benchmark $\uparrow$} \\
\cmidrule(lr){2-4} \cmidrule(lr){5-6}
Setting & Prompt & Judge & Evolution & Physics-IQ & StEvo \\
\midrule
Structured & Max & -- & -- & 44.9 & 51.6 \\
Ours & Max & Max & DS-V4 & \textbf{48.3} & \textbf{67.4} \\
\midrule
\multicolumn{6}{l}{\textit{Replace one agent}} \\
Prompt & \underline{27B} & Max & DS-V4 & 44.6 & 59.3 \\
Judge & Max & \underline{27B} & DS-V4 & 43.9 & 65.6 \\
Evolution & Max & Max & \underline{Max} & 46.5 & 64.7 \\
\bottomrule
\end{tabular}
\end{table}

\textbf{Ablation on agent type.} Tab.~\ref{tab:ablation_agent_type} replaces one agent at a time during harness evolution on MiniMax-H3, keeping everything else fixed. Every replacement lowers both benchmarks, so all three agents contribute to the final harness. The prompt agent and the judge agent matter most. A smaller prompt agent loses 8.1 points on StEvo, and a smaller judge loses 4.4 points on Physics-IQ. With either change, Physics-IQ even falls below the structured prompt, while StEvo stays 7.7 to 14.0 points above it. A likely reason is that Physics-IQ checks fine motion and contact, which needs precise prompts and feedback, while StEvo checks coarser state changes. The evolution agent matters least: replacing it with Qwen3.8-Max loses only 1.8 and 2.7 points, and the harness still beats every baseline on MiniMax-H3. Our method thus needs reliable prompt writing and judging, but not a specific evolution agent.

\section{Conclusion}
\label{sec:conclusion}

We studied how reasoning agents can better convey physical knowledge to video diffusion models (VDMs). FantasyEvo uses feedback from each VDM to learn a harness that guides the agent in writing prompts for that model. The harness is frozen at test time, so each task takes one agent call and one VDM call. FantasyEvo improves over raw prompting on both benchmarks across all three VDMs. We hope this work encourages further research on how reasoning agents communicate with video generation models.

\clearpage
\newpage

\bibliographystyle{unsrt}
\bibliography{references}

@String(CVPR= {IEEE Conf. Comput. Vis. Pattern Recog.})

@String(ICCV= {Int. Conf. Comput. Vis.})

@String(AAAI = {AAAI})

@String(CVPRW= {IEEE Conf. Comput. Vis. Pattern Recog. Worksh.})

@String(CVPR  = {CVPR})

@String(ICCV  = {ICCV})

@String(CVPRW= {CVPRW})

@article{brooks2024video,
  title={Video generation models as world simulators},
  author={Brooks, Tim and Peebles, Bill and Holmes, Connor and DePue, Will and Guo, Yufei and Jing, Leo and Schnurr, David and Taylor, Joe and Luhman, Troy and Luhman, Eric and others},
  journal={OpenAI Blog},
  volume={1},
  number={8},
  pages={1},
  year={2024}
}

@article{kong2024hunyuanvideo,
  title={Hunyuanvideo: A systematic framework for large video generative models},
  author={Kong, Weijie and Tian, Qi and Zhang, Zijian and Min, Rox and Dai, Zuozhuo and Zhou, Jin and Xiong, Jiangfeng and Li, Xin and Wu, Bo and Zhang, Jianwei and others},
  journal={arXiv preprint arXiv:2412.03603},
  year={2024}
}

@article{wan2025wan,
  title={Wan: Open and advanced large-scale video generative models},
  author={Wan, Team and Wang, Ang and Ai, Baole and Wen, Bin and Mao, Chaojie and Xie, Chen-Wei and Chen, Di and Yu, Feiwu and Zhao, Haiming and Yang, Jianxiao and others},
  journal={arXiv preprint arXiv:2503.20314},
  year={2025}
}

@inproceedings{motamed2026generative,
  title={Do generative video models understand physical principles?},
  author={Motamed, Saman and Culp, Laura and Swersky, Kevin and Jaini, Priyank and Geirhos, Robert},
  booktitle={2026 IEEE/CVF Winter Conference on Applications of Computer Vision (WACV)},
  pages={948--958},
  year={2026},
  organization={IEEE}
}

@article{meng2024towards,
  title={Towards world simulator: Crafting physical commonsense-based benchmark for video generation},
  author={Meng, Fanqing and Liao, Jiaqi and Tan, Xinyu and Shao, Wenqi and Lu, Quanfeng and Zhang, Kaipeng and Cheng, Yu and Li, Dianqi and Qiao, Yu and Luo, Ping},
  journal={arXiv preprint arXiv:2410.05363},
  year={2024}
}

@inproceedings{ma2026out,
  title={Out of sight, out of mind? evaluating state evolution in video world models},
  author={Ma, Ziqi and Liufu, Mengzhan and Gkioxari, Georgia},
  booktitle={European Conference on Computer Vision},
  pages={129--146},
  year={2026},
  organization={Springer}
}

@inproceedings{bansal2026videophy,
  title={Videophy-2: A challenging action-centric physical commonsense evaluation in video generation},
  author={Bansal, Hritik and Peng, Clark and Bitton, Yonatan and Goldenberg, Roman and Grover, Aditya and Chang, Kai-Wei},
  booktitle={International Conference on Learning Representations},
  volume={2026},
  pages={118456--118470},
  year={2026}
}

@inproceedings{bansal2025videophy,
  title={Videophy: Evaluating physical commonsense for video generation},
  author={Bansal, Hritik and Lin, Zongyu and Xie, Tianyi and Zong, Zeshun and Yarom, Michal and Bitton, Yonatan and Jiang, Chenfanfu and Sun, Yizhou and Chang, Kai-Wei and Grover, Aditya},
  booktitle={International Conference on Learning Representations},
  volume={2025},
  pages={102075--102121},
  year={2025}
}

@article{zhou2026tide,
  title={TIDE: Trajectory-Informed Skill Distillation for Efficient Text-to-Video Prompt Optimization},
  author={Zhou, Chengxing and Huang, Yuliang and Song, Jingwei},
  journal={arXiv preprint arXiv:2610.15098},
  year={2026}
}

@inproceedings{xue2025phyt2v,
  title={Phyt2v: Llm-guided iterative self-refinement for physics-grounded text-to-video generation},
  author={Xue, Qiyao and Yin, Xiangyu and Yang, Boyuan and Gao, Wei},
  booktitle={2025 IEEE/CVF Conference on Computer Vision and Pattern Recognition (CVPR)},
  pages={18826--18836},
  year={2025},
  organization={IEEE}
}

@article{miao2026video,
  title={Do video generators track the world across segments? a benchmark and method for world-state reasoning in video continuation},
  author={Miao, Yingmao and Zhang, Pengfei and Xu, Chaoran and Yu, Meng and Tang, Jing and Chu, Xiangxiang and Shen, Chao and Lin, Chenhao},
  journal={arXiv preprint arXiv:2609.03673},
  year={2026}
}

@inproceedings{lian2024llm,
  title={Llm-grounded video diffusion models},
  author={Lian, Long and Shi, Baifeng and Yala, Adam and Li, Boyi and others},
  booktitle={International Conference on Learning Representations},
  volume={2024},
  pages={50207--50227},
  year={2024}
}

@article{lin2023videodirectorgpt,
  title={Videodirectorgpt: Consistent multi-scene video generation via llm-guided planning},
  author={Lin, Han and Zala, Abhay and Cho, Jaemin and Bansal, Mohit},
  journal={arXiv preprint arXiv:2309.15091},
  year={2023}
}

@article{tang2024worldcoder,
  title={Worldcoder, a model-based llm agent: Building world models by writing code and interacting with the environment},
  author={Tang, Hao and Key, Darren and Ellis, Kevin},
  journal={Advances in Neural Information Processing Systems},
  volume={37},
  pages={70148--70212},
  year={2024}
}

@article{dainese2024generating,
  title={Generating code world models with large language models guided by monte carlo tree search},
  author={Dainese, Nicola and Merler, Matteo and Alakuijala, Minttu and Marttinen, Pekka},
  journal={Advances in Neural Information Processing Systems},
  volume={37},
  pages={60429--60474},
  year={2024}
}

@article{zhang2026code2worlds,
  title={Code2worlds: Empowering coding llms for 4d world generation},
  author={Zhang, Yi and Wang, Yunshuang and Zhang, Zeyu and Tang, Hao},
  journal={arXiv preprint arXiv:2602.11757},
  year={2026}
}

@article{chen2026code,
  title={Code world model: Coding agent as world brain},
  author={Chen, Yiwen and Lin, Guosheng and Zhang, Chi},
  journal={arXiv preprint arXiv:2608.25927},
  year={2026}
}

@article{li2026recursive,
  title={Recursive code world models: Building complex worlds through recursive scene programs},
  author={Li, Zhiqi and Liao, Yuxuan and Zhu, Bo},
  journal={arXiv preprint arXiv:2609.11499},
  year={2026}
}

@inproceedings{lv2024gpt4motion,
  title={Gpt4motion: Scripting physical motions in text-to-video generation via blender-oriented gpt planning},
  author={Lv, Jiaxi and Huang, Yi and Yan, Mingfu and Huang, Jiancheng and Liu, Jianzhuang and Liu, Yifan and Wen, Yafei and Chen, Xiaoxin and Chen, Shifeng},
  booktitle={2024 IEEE/CVF Conference on Computer Vision and Pattern Recognition Workshops (CVPRW)},
  pages={1430--1440},
  year={2024},
  organization={IEEE}
}

@inproceedings{liu2024physgen,
  title={Physgen: Rigid-body physics-grounded image-to-video generation},
  author={Liu, Shaowei and Ren, Zhongzheng and Gupta, Saurabh and Wang, Shenlong},
  booktitle={European Conference on Computer Vision},
  pages={360--378},
  year={2024},
  organization={Springer}
}

@article{feng2026newton,
  title={NEWTON: Agentic Planning for Physically Grounded Video Generation},
  author={Feng, Yuxiang and Wang, Juncheng and Xu, Chao and Qian, Yijie and Wang, Huihan and Hou, Wenlong and Liu, Yang and Sun, Baigui and Liu, Yong and Wang, Shujun},
  journal={arXiv preprint arXiv:2605.18396},
  year={2026}
}

@article{seedance,
  title   = {Seedance 2.0: Advancing Video Generation for World Complexity},
  author  = {Seedance, Team and Chen, De and Chen, Liyang and Chen, Xin and Chen, Ying and Chen, Zhuo and Chen, Zhuowei and Cheng, Feng and Cheng, Tianheng and Cheng, Yufeng and others},
  journal = {arXiv preprint arXiv:2604.14148},
  year    = {2026}
}

@misc{kling,
  title  = {{KLING AI}},
  author = {{KlingAI}},
  year   = {2024},
  url    = {https://www.klingai.com/}
}

@misc{minimaxh3,
  title  = {{MiniMax H3}: An Open Model Breaking the Boundaries between Tasks and Modalities},
  author = {{MiniMax}},
  year   = {2026},
  url    = {https://www.minimax.io/blog/minimax-h3}
}

@techreport{veo3,
  title       = {Veo: A Text-to-Video Generation System},
  author      = {{Google DeepMind}},
  institution = {Google DeepMind},
  year        = {2025},
  url         = {https://storage.googleapis.com/deepmind-media/veo/Veo-3-Tech-Report.pdf}
}

@inproceedings{physicsiq,
  title = {{Do Generative Video Models Understand Physical Principles?}},
  author = {Motamed, Saman and Culp, Laura and Swersky, Kevin and Jaini, Priyank and Geirhos, Robert},
  booktitle = {Proceedings of the IEEE/CVF Winter Conference on Applications of Computer Vision (WACV)},
  year = {2026},
  pages = {948--958},
  month = {March},
  url = {https://openaccess.thecvf.com/content/WACV2026/html/Motamed_Do_Generative_Video_Models_Understand_Physical_Principles_WACV_2026_paper.html}
}

@inproceedings{kang2024howfar,
  title = {{How Far Is Video Generation from World Model: A Physical Law Perspective}},
  author = {Kang, Bingyi and Yue, Yang and Lu, Rui and Lin, Zhijie and Zhao, Yang and Wang, Kaixin and Huang, Gao and Feng, Jiashi},
  booktitle = {Proceedings of the 42nd International Conference on Machine Learning},
  year = {2025},
  pages = {28991--29017},
  volume = {267},
  series = {Proceedings of Machine Learning Research},
  publisher = {PMLR},
  url = {https://proceedings.mlr.press/v267/kang25g.html}
}

@inproceedings{worldmodelbench,
  title = {{WorldModelBench: Judging Video Generation Models As World Models}},
  author = {Li, Dacheng and Fang, Yunhao and Chen, Yukang and Yang, Shuo and Cao, Shiyi and Wong, Justin and Luo, Michael and Wang, Xiaolong and Yin, Hongxu and Gonzalez, Joseph E. and Stoica, Ion and Han, Song and Lu, Yao},
  booktitle = {Advances in Neural Information Processing Systems},
  year = {2025},
  volume = {38},
  doi = {10.52202/085713-1834},
  url = {https://papers.neurips.cc/paper_files/paper/2025/hash/4ec03ed08a3fcb59e1c815b5598beff1-Abstract-Datasets_and_Benchmarks_Track.html}
}

@article{vbench2,
  title = {{VBench-2.0: Advancing Video Generation Benchmark Suite for Intrinsic Faithfulness}},
  author = {Zheng, Dian and Huang, Ziqi and Liu, Hongbo and Zou, Kai and He, Yinan and Zhang, Fan and Gu, Lulu and Zhang, Yuanhan and He, Jingwen and Zheng, Wei-Shi and Qiao, Yu and Liu, Ziwei},
  journal = {arXiv preprint arXiv:2503.21755},
  year = {2025},
  url = {https://arxiv.org/abs/2503.21755}
}

@inproceedings{physdreamer,
  title = {{PhysDreamer: Physics-Based Interaction with 3D Objects via Video Generation}},
  author = {Zhang, Tianyuan and Yu, Hong-Xing and Wu, Rundi and Feng, Brandon Y. and Zheng, Changxi and Snavely, Noah and Wu, Jiajun and Freeman, William T.},
  booktitle = {European Conference on Computer Vision},
  year = {2024},
  organization = {Springer},
  url = {https://physdreamer.github.io/}
}

@inproceedings{motioncraft,
  title = {{MotionCraft: Physics-Based Zero-Shot Video Generation}},
  author = {Montanaro, Antonio and Savant Aira, Luca and Aiello, Emanuele and Valsesia, Diego and Magli, Enrico},
  booktitle = {Advances in Neural Information Processing Systems},
  year = {2024},
  volume = {37},
  doi = {10.52202/079017-3914},
  url = {https://proceedings.neurips.cc/paper_files/paper/2024/hash/ded49449e9239a47057c33ef8d42c26e-Abstract-Conference.html}
}

@inproceedings{wisa,
  title = {{WISA: World simulator assistant for physics-aware text-to-video generation}},
  author = {Wang, Jing and Ma, Ao and Cao, Ke and Zheng, Jun and Feng, Jiasong and Zhang, Zhanjie and Pang, Wanyuan and Liang, Xiaodan},
  booktitle = {Advances in Neural Information Processing Systems},
  year = {2025},
  volume = {38},
  doi = {10.52202/085713-0191},
  url = {https://proceedings.neurips.cc/paper_files/paper/2025/hash/0856bc553d3e3b9827e5140d0ad3bf8d-Abstract-Conference.html}
}

@article{physmaster,
  title = {{PhysMaster: Mastering Physical Representation for Video Generation via Reinforcement Learning}},
  author = {Ji, Sihui and Chen, Xi and Tao, Xin and Wan, Pengfei and Zhao, Hengshuang},
  journal = {arXiv preprint arXiv:2510.13809},
  year = {2025},
  url = {https://arxiv.org/abs/2510.13809}
}

@inproceedings{videorepa,
  title = {{VideoREPA: Learning Physics for Video Generation through Relational Alignment with Foundation Models}},
  author = {Zhang, Xiangdong and Liao, Jiaqi and Zhang, Shaofeng and Meng, Fanqing and Wan, Xiangpeng and Yan, Junchi and Cheng, Yu},
  booktitle = {Advances in Neural Information Processing Systems},
  year = {2025},
  volume = {38},
  doi = {10.52202/085713-4091},
  url = {https://proceedings.neurips.cc/paper_files/paper/2025/hash/b1d4973f5d21708abef3cd6f17d842c8-Abstract-Conference.html}
}

@inproceedings{forceprompting,
  title = {{Force Prompting: Video Generation Models Can Learn And Generalize Physics-based Control Signals}},
  author = {Gillman, Nate and Herrmann, Charles and Freeman, Michael and Aggarwal, Daksh and Luo, Evan and Sun, Deqing and Sun, Chen},
  booktitle = {Advances in Neural Information Processing Systems},
  year = {2025},
  volume = {38},
  doi = {10.52202/085713-3448},
  url = {https://proceedings.neurips.cc/paper_files/paper/2025/hash/953dcf20cbd275465066ad63ea111f13-Abstract-Conference.html}
}

@inproceedings{freebloom,
  title = {{Free-Bloom: Zero-Shot Text-to-Video Generator with LLM Director and LDM Animator}},
  author = {Huang, Hanzhuo and Feng, Yufan and Shi, Cheng and Xu, Lan and Yu, Jingyi and Yang, Sibei},
  booktitle = {Advances in Neural Information Processing Systems},
  year = {2023},
  volume = {36},
  url = {https://proceedings.neurips.cc/paper_files/paper/2023/hash/52f050499cf82fa8efb588e263f6f3a7-Abstract.html}
}

@inproceedings{vlogger,
  title = {{Vlogger: Make Your Dream A Vlog}},
  author = {Zhuang, Shaobin and Li, Kunchang and Chen, Xinyuan and Wang, Yaohui and Liu, Ziwei and Qiao, Yu and Wang, Yali},
  booktitle = {Proceedings of the IEEE/CVF Conference on Computer Vision and Pattern Recognition (CVPR)},
  year = {2024},
  pages = {8806--8817},
  month = {June},
  url = {https://openaccess.thecvf.com/content/CVPR2024/html/Zhuang_Vlogger_Make_Your_Dream_A_Vlog_CVPR_2024_paper.html}
}

@article{dreamfactory,
  title = {{DreamFactory: Pioneering Multi-Scene Long Video Generation with a Multi-Agent Framework}},
  author = {Xie, Zhifei and Tang, Daniel and Tan, Dingwei and Klein, Jacques and Bissyand{\'e}, Tegawend{\'e} F. and Ezzini, Saad},
  journal = {arXiv preprint arXiv:2408.11788},
  year = {2024},
  url = {https://arxiv.org/abs/2408.11788}
}

@article{diffphy,
  title = {{Think Before You Diffuse: Infusing Physical Rules into Video Diffusion}},
  author = {Zhang, Ke and Xiao, Cihan and Xu, Jiacong and Mei, Yiqun and Patel, Vishal M.},
  journal = {arXiv preprint arXiv:2505.21653},
  year = {2025},
  url = {https://arxiv.org/abs/2505.21653}
}

@inproceedings{ape,
  title = {{Large Language Models are Human-Level Prompt Engineers}},
  author = {Yongchao Zhou and Andrei Ioan Muresanu and Ziwen Han and Keiran Paster and Silviu Pitis and Harris Chan and Jimmy Ba},
  booktitle = {International Conference on Learning Representations},
  year = {2023},
  url = {https://openreview.net/forum?id=92gvk82DE-}
}

@inproceedings{protegi,
  title = {{Automatic Prompt Optimization with ``Gradient Descent'' and Beam Search}},
  author = {Pryzant, Reid and Iter, Dan and Li, Jerry and Lee, Yin and Zhu, Chenguang and Zeng, Michael},
  booktitle = {Proceedings of the 2023 Conference on Empirical Methods in Natural Language Processing},
  year = {2023},
  pages = {7957--7968},
  publisher = {Association for Computational Linguistics},
  doi = {10.18653/v1/2023.emnlp-main.494},
  url = {https://aclanthology.org/2023.emnlp-main.494/}
}

@inproceedings{promptagent,
  title = {{PromptAgent: Strategic Planning with Language Models Enables Expert-level Prompt Optimization}},
  author = {Wang, Xinyuan and Li, Chenxi and Wang, Zhen and Bai, Fan and Luo, Haotian and Zhang, Jiayou and Jojic, Nebojsa and Xing, Eric and Hu, Zhiting},
  booktitle = {International Conference on Learning Representations},
  year = {2024},
  volume = {2024},
  pages = {23967--24001},
  url = {https://proceedings.iclr.cc/paper_files/paper/2024/file/686a3f32067838c8dbb68da6e9e3cf69-Paper-Conference.pdf}
}

@inproceedings{opro,
  title = {{Large Language Models as Optimizers}},
  author = {Yang, Chengrun and Wang, Xuezhi and Lu, Yifeng and Liu, Hanxiao and Le, Quoc V and Zhou, Denny and Chen, Xinyun},
  booktitle = {International Conference on Learning Representations},
  year = {2024},
  volume = {2024},
  pages = {12028--12068},
  url = {https://proceedings.iclr.cc/paper_files/paper/2024/file/3339f19c5fcee3ad74502947a32be9e6-Paper-Conference.pdf}
}

@inproceedings{evoprompt,
  title = {{Connecting Large Language Models with Evolutionary Algorithms Yields Powerful Prompt Optimizers}},
  author = {Guo, Qingyan and Wang, Rui and Guo, Junliang and Li, Bei and Song, Kaitao and Tan, Xu and Liu, Guoqing and Bian, Jiang and Yang, Yujiu},
  booktitle = {International Conference on Learning Representations},
  year = {2024},
  volume = {2024},
  pages = {34133--34156},
  url = {https://proceedings.iclr.cc/paper_files/paper/2024/file/9156b0f6dfa9bbd18c79cc459ef5d61c-Paper-Conference.pdf}
}

@inproceedings{promptbreeder,
  title = {{Promptbreeder: Self-Referential Self-Improvement via Prompt Evolution}},
  author = {Fernando, Chrisantha and Banarse, Dylan Sunil and Michalewski, Henryk and Osindero, Simon and Rockt\"{a}schel, Tim},
  booktitle = {Proceedings of the 41st International Conference on Machine Learning},
  year = {2024},
  volume = {235},
  pages = {13481--13544},
  publisher = {PMLR},
  url = {https://proceedings.mlr.press/v235/fernando24a.html}
}

@inproceedings{gepa,
  title = {{GEPA: Reflective Prompt Evolution Can Outperform Reinforcement Learning}},
  author = {Agrawal, Lakshya A and Tan, Shangyin and Soylu, Dilara and Ziems, Noah and Khare, Rishi and Opsahl-Ong, Krista and Singhvi, Arnav and Shandilya, Herumb and Ryan, Michael J and Jiang, Meng and Potts, Christopher and Sen, Koushik and Dimakis, Alex and Stoica, Ion and Klein, Dan and Zaharia, Matei and Khattab, Omar},
  booktitle = {International Conference on Learning Representations},
  year = {2026},
  volume = {2026},
  pages = {8479--8565},
  url = {https://proceedings.iclr.cc/paper_files/paper/2026/file/0e9e708b6f48e14fd0ac29e167413f76-Paper-Conference.pdf}
}

@article{textgrad,
  title = {{Optimizing generative AI by backpropagating language model feedback}},
  author = {Yuksekgonul, Mert and Bianchi, Federico and Boen, Joseph and Liu, Sheng and Lu, Pan and Huang, Zhi and Guestrin, Carlos and Zou, James},
  journal = {Nature},
  year = {2025},
  volume = {639},
  number = {8055},
  pages = {609--616},
  publisher = {Springer Science and Business Media LLC},
  doi = {10.1038/s41586-025-08661-4},
  url = {https://doi.org/10.1038/s41586-025-08661-4}
}

@inproceedings{dspy,
  title = {{DSPy: Compiling Declarative Language Model Calls into State-of-the-Art Pipelines}},
  author = {Khattab, Omar and Singhvi, Arnav and Maheshwari, Paridhi and Zhang, Zhiyuan and Santhanam, Keshav and A, Sri Vardhamanan and Haq, Saiful and Sharma, Ashutosh and Joshi, Thomas and Moazam, Hanna and Miller, Heather and Zaharia, Matei and Potts, Christopher},
  booktitle = {International Conference on Learning Representations},
  year = {2024},
  volume = {2024},
  pages = {54928--54958},
  url = {https://proceedings.iclr.cc/paper_files/paper/2024/file/f1cf02ce09757f57c3b93c0db83181e0-Paper-Conference.pdf}
}

@inproceedings{mipro,
  title = {{Optimizing Instructions and Demonstrations for Multi-Stage Language Model Programs}},
  author = {Opsahl-Ong, Krista and Ryan, Michael J and Purtell, Josh and Broman, David and Potts, Christopher and Zaharia, Matei and Khattab, Omar},
  booktitle = {Proceedings of the 2024 Conference on Empirical Methods in Natural Language Processing},
  year = {2024},
  pages = {9340--9366},
  publisher = {Association for Computational Linguistics},
  doi = {10.18653/v1/2024.emnlp-main.525},
  url = {https://aclanthology.org/2024.emnlp-main.525/}
}

@inproceedings{reflexion,
  title = {{Reflexion: language agents with verbal reinforcement learning}},
  author = {Shinn, Noah and Cassano, Federico and Gopinath, Ashwin and Narasimhan, Karthik and Yao, Shunyu},
  booktitle = {Advances in Neural Information Processing Systems},
  year = {2023},
  volume = {36},
  pages = {8634--8652},
  publisher = {Curran Associates, Inc.},
  doi = {10.52202/075280-0377},
  url = {https://proceedings.neurips.cc/paper_files/paper/2023/file/1b44b878bb782e6954cd888628510e90-Paper-Conference.pdf}
}

@article{expel,
  title = {{ExpeL: LLM Agents Are Experiential Learners}},
  author = {Zhao, Andrew and Huang, Daniel and Xu, Quentin and Lin, Matthieu and Liu, Yong-Jin and Huang, Gao},
  journal = {Proceedings of the AAAI Conference on Artificial Intelligence},
  year = {2024},
  volume = {38},
  number = {17},
  pages = {19632--19642},
  publisher = {Association for the Advancement of Artificial Intelligence (AAAI)},
  doi = {10.1609/aaai.v38i17.29936},
  url = {https://doi.org/10.1609/aaai.v38i17.29936}
}

@article{voyager,
  title = {{Voyager: An Open-Ended Embodied Agent with Large Language Models}},
  author = {Guanzhi Wang and Yuqi Xie and Yunfan Jiang and Ajay Mandlekar and Chaowei Xiao and Yuke Zhu and Linxi Fan and Anima Anandkumar},
  journal = {Transactions on Machine Learning Research},
  year = {2024},
  url = {https://openreview.net/forum?id=ehfRiF0R3a}
}

@inproceedings{awm,
  title = {{Agent Workflow Memory}},
  author = {Wang, Zora Zhiruo and Mao, Jiayuan and Fried, Daniel and Neubig, Graham},
  booktitle = {Proceedings of the 42nd International Conference on Machine Learning},
  year = {2025},
  volume = {267},
  pages = {63897--63911},
  publisher = {PMLR},
  url = {https://proceedings.mlr.press/v267/wang25bx.html}
}

@inproceedings{dynamiccheatsheet,
  title = {{Dynamic Cheatsheet: Test-Time Learning with Adaptive Memory}},
  author = {Suzgun, Mirac and Yuksekgonul, Mert and Bianchi, Federico and Jurafsky, Dan and Zou, James},
  booktitle = {Proceedings of the 19th Conference of the {E}uropean Chapter of the {A}ssociation for {C}omputational {L}inguistics (Volume 1: Long Papers)},
  year = {2026},
  pages = {7080--7106},
  publisher = {Association for Computational Linguistics},
  doi = {10.18653/v1/2026.eacl-long.333},
  url = {https://aclanthology.org/2026.eacl-long.333/}
}

@inproceedings{ace,
  title = {{Agentic Context Engineering: Evolving Contexts for Self-Improving Language Models}},
  author = {Zhang, Qizheng and Hu, Changran and Upasani, Shubhangi and Ma, Boyuan and Hong, Fenglu and Kamanuru, Vamsidhar and Rainton, Jay and Wu, Chen and Ji, Mengmeng and Li, Hanchen and Thakker, Urmish and Zou, James Y and Olukotun, Kunle},
  booktitle = {International Conference on Learning Representations},
  year = {2026},
  volume = {2026},
  pages = {86069--86100},
  url = {https://proceedings.iclr.cc/paper_files/paper/2026/file/8a94ff6f922d995d7d3f4ebf4143e442-Paper-Conference.pdf}
}

@inproceedings{promptist,
  title = {{Optimizing Prompts for Text-to-Image Generation}},
  author = {Hao, Yaru and Chi, Zewen and Dong, Li and Wei, Furu},
  booktitle = {Advances in Neural Information Processing Systems},
  year = {2023},
  volume = {36},
  url = {https://proceedings.neurips.cc/paper_files/paper/2023/hash/d346d91999074dd8d6073d4c3b13733b-Abstract.html}
}

@inproceedings{promptavideo,
  title = {{Prompt-A-Video: Prompt Your Video Diffusion Model via Preference-Aligned LLM}},
  author = {Ji, Yatai and Zhang, Jiacheng and Wu, Jie and Zhang, Shilong and Chen, Shoufa and Ge, Chongjian and Sun, Peize and Chen, Weifeng and Shao, Wenqi and Xiao, Xuefeng and Huang, Weilin and Luo, Ping},
  booktitle = {Proceedings of the IEEE/CVF International Conference on Computer Vision (ICCV)},
  year = {2025},
  pages = {18725--18735},
  month = {October},
  url = {https://openaccess.thecvf.com/content/ICCV2025/html/Ji_Prompt-A-Video_Prompt_Your_Video_Diffusion_Model_via_Preference-Aligned_LLM_ICCV_2025_paper.html}
}

@inproceedings{vpo,
  title = {{VPO: Aligning Text-to-Video Generation Models with Prompt Optimization}},
  author = {Cheng, Jiale and Lyu, Ruiliang and Gu, Xiaotao and Liu, Xiao and Xu, Jiazheng and Lu, Yida and Teng, Jiayan and Yang, Zhuoyi and Dong, Yuxiao and Tang, Jie and Wang, Hongning and Huang, Minlie},
  booktitle = {Proceedings of the IEEE/CVF International Conference on Computer Vision (ICCV)},
  year = {2025},
  pages = {15636--15645},
  month = {October},
  url = {https://openaccess.thecvf.com/content/ICCV2025/html/Cheng_VPO_Aligning_Text-to-Video_Generation_Models_with_Prompt_Optimization_ICCV_2025_paper.html}
}

@inproceedings{rapo,
  title = {{The Devil is in the Prompts: Retrieval-Augmented Prompt Optimization for Text-to-Video Generation}},
  author = {Gao, Bingjie and Gao, Xinyu and Wu, Xiaoxue and Zhou, Yujie and Qiao, Yu and Niu, Li and Chen, Xinyuan and Wang, Yaohui},
  booktitle = {Proceedings of the IEEE/CVF Conference on Computer Vision and Pattern Recognition (CVPR)},
  year = {2025},
  pages = {3173--3183},
  month = {June},
  url = {https://openaccess.thecvf.com/content/CVPR2025/html/Gao_The_Devil_is_in_the_Prompts_Retrieval-Augmented_Prompt_Optimization_for_CVPR_2025_paper.html}
}

@article{zhao2026qwen,
  title={Qwen-image-2.0 technical report},
  author={Zhao, Bing and Wu, Chenfei and Li, Deqing and Meng, Hao and Li, Jiahao and Zhang, Jie and Zhou, Jingren and Lin, Junyang and Gao, Kaiyuan and Cao, Kuan and others},
  journal={arXiv preprint arXiv:2605.10730},
  year={2026}
}

@misc{qwen38,
  title  = {{Qwen3.8-Max}: A New Bar for Coding and Cowork},
  author = {{Qwen Team}},
  month  = {August},
  year   = {2026},
  url    = {https://qwen.ai/blog?id=qwen3.8}
}

@article{agarwal2026cosmos,
  title={Cosmos 3: Omnimodal world models for physical ai},
  author={Agarwal, Niket and Ali, Arslan and Allen, Jon and Antolini, Martin and Aubame, Adeline and Azzolini, Alisson and Bai, Junjie and Bala, Maciej and Balaji, Yogesh and Bapst, Josh and others},
  journal={arXiv preprint arXiv:2606.02800},
  year={2026}
}

@article{xu2026deepseek,
  title={Deepseek-v4: Towards highly efficient million-token context intelligence},
  author={Xu, Anyi and Lin, Bangcai and Xue, Bing and Wang, Bingxuan and Xu, Bingzheng and Wu, Bochao and Zhang, Bowei and Lin, Chaofan and Dong, Chen and Ling, Chenchen and others},
  journal={arXiv preprint arXiv:2606.19348},
  year={2026}
}

\end{document}